\documentclass{article}

     \usepackage[preprint,nonatbib]{neurips_2020}

\usepackage[utf8]{inputenc} 
\usepackage[T1]{fontenc}    
\usepackage{hyperref}
\usepackage{cleveref}
\usepackage{url}            
\usepackage{booktabs}       
\usepackage{amsfonts}       
\usepackage{nicefrac}       
\usepackage{microtype}      

\usepackage{amsmath}
\usepackage{amssymb}
\usepackage{graphicx}
\usepackage{multirow}
\usepackage{times}  
\usepackage{helvet} 
\usepackage{courier}  

\usepackage{algorithm} 
\usepackage{algorithmic}

\DeclareMathOperator*{\argmax}{arg\,max}

\title{Decorrelated Double Q-learning}

\author{%
  Gang Chen \\
  Department of Computer Science\\
  SUNY at Buffalo, \\
  Buffalo, NY 14260 \\
  \texttt{gangchen@buffalo.edu} \\
}

\begin{document}

\maketitle

\begin{abstract}
Q-learning with value function approximation may have the poor performance because of overestimation bias and imprecise estimate. Specifically, overestimation bias is from the maximum operator over noise estimate, which is exaggerated using the estimate of a subsequent state. Inspired by the recent advance of deep reinforcement learning and Double Q-learning, we introduce the decorrelated double Q-learning (D2Q). Specifically, we introduce the decorrelated regularization item to reduce the correlation between value function approximators, which can lead to less biased estimation and low variance. 
The experimental results on a suite of MuJoCo continuous control tasks demonstrate that our decorrelated double Q-learning can effectively improve the performance.
\end{abstract}

\section{Introduction}

Q-learning \cite{Watkins92} as a model free reinforcement learning approach has gained popularity, especially under the advance of deep neural networks \cite{Mnih13}. In general, it combines the neural network approximators with the actor-critic architectures \cite{Witten77,Konda00}, which has an actor network to control how the agent behaves and a critic to evaluate how good the action taken is. 

The Deep Q-Network (DQN) algorithm \cite{Mnih13} firstly applied the deep neural network to approximate the action-value function in Q-learning and shown remarkably good and stable results by introducing a target network and Experience Replay buffer to stabilize the training. Lillicrap et al. extends Q-learning under deterministic policy gradient \cite{Silver14} to handle continuous action space \cite{LillicrapHPHETS15} with target networks. Except the training stability, another issue Q-learning suffered is overestimation bias, which was first investigated in \cite{Thrun93}. Because of the noise in function approximation, the maximum operator in Q-learning can lead to overestimation of state-action values. And, the overestimation property is also observed in deterministic continuous policy control \cite{Silver14}. In particular, with the imprecise function approximation, the maximization of a noisy value will induce overestimation to the action value function. This inaccuracy could be even worse (e.g. error accumulation) under temporal difference learning \cite{SuttonB98}, in which bootstrapping method is used to update the value function using the estimate of a subsequent state. 

Given overestimation bias caused by maximum operator of noise estimate, many methods have been proposed to address this issue. Double Q-learning \cite{Hasselt10} mitigates the overestimation effect by introducing two independently critics to estimate the maximum value of a set of stochastic values. Averaged-DQN \cite{AnschelBS17} takes the average of previously learned Q-values estimates, which results in a more stable training procedure, as well as reduces approximation error variance in the target values. Recently, Twin Delayed Deep Deterministic Policy Gradients (TD3) \cite{FujimotoHM18} extends the Double Q-learning, by using the minimum of two critics to limit the overestimated bias in actor-critic network. A soft Q-learning algorithm \cite{HaarnojaZAL18}, called soft actor-critic, leverages the similar strategy as TD3, while including the maximum entropy to balance exploration and exploitation. 
Maxmin Q-learning \cite{Lan20} proposes the use of an ensembling scheme to handle overestimation bias in Q-Learning. 

This work suggests a different solution to the overestimation phenomena, called decorrelated double Q-learning, based on reducing the noise estimate in Q-values. On the one hand, we want to make the two value function approximators as independent as possible to mitigate overestimation bias. On the other hand, we should reduce the variance caused by imprecise estimate. Our decorrelated double Q-learning proposes an objective function to minimize 
the correlation of two critics, and meanwhile reduces the target approximation error variance with control variate methods. 
Finally, we provide experimental results on MuJoCo games and show significant improvement compared to competitive baselines.
%

\section{Background}
In this section, we introduce the reinforcement learning problems, as well as notions that will be used in the following sections.
\subsection{Problem settings and Notations}
We consider the model-free reinforcement learning problem (i.e. optimal policy existed) with sequential interactions between an agent and its environment \cite{SuttonB98} in order to maximize a cumulative return. At every time step $t$, the agent selects an action $a_t$ in the state $s_t$ according its policy and receives a scalar reward $r_t(s_t, a_t)$, and then transit to the next state $s_{t+1}$. The problem is modeled as Markov decision process (MDP) with tuple: $(\mathcal{S}, \mathcal{A}, p(s_{0}), p(s_{t+1}|s_t, a_t),  r(s_t, a_t), \gamma )$. Here, $\mathcal{S}$ and $\mathcal{A}$ indicate the state and action space respectively, $p(s_{0})$ is the initial state distribution. $p(s_{t+1}|s_t, a_t)$ is the state transition to $s_{t+1}$ given the current state $s_t$ and action $a_t$, $r(s_t, a_t)$ is reward from the environment after the agent taking action $a_t$ in state $s_t$ and $\gamma$ is discount factor, which is necessary to decay the future rewards ensuring finite returns. We model the agent's behavior with 
$\pi_{\theta} (a|s)$, which is a parametric distribution from a neural network. 

Suppose we have the finite length trajectory while the agent interacting with the environment. The return under the policy $\pi$ for a trajectory $\tau = {  ( s_t,  a_t   )  }_{t=0}^T$   
\begin{align}\label{eq:obj}
J(\theta) & = \mathbb{E}_{\tau \sim \pi_{\theta}(\tau)} [  r(\tau)  ] = \mathbb{E}_{\tau \sim \pi_{\theta}(\tau)} [  R_0^T ] \nonumber \\ 
 & = \mathbb{E}_{\tau \sim \pi_{\theta}(\tau)} [  \sum_{t=0}^T  \gamma^t r(  s_t, a_t   )   ] 
\end{align}
where $\pi_{\theta}(\tau)$ denotes the distribution of trajectories, 
\begin{align}
 p(\tau) &= \pi(s_0, a_0, s_1, ..., s_T, a_T) \nonumber \\
 & = p(s_0) \prod_{t=0}^T \pi_{\theta}(a_t | s_t) p(s_{t+1} | s_t, a_t) 
\end{align}
The goal of reinforcement learning is to learn a policy $\pi$ which can maximize the expected returns
\begin{equation} \label{eq:eq1}
\theta = \argmax_{\theta} J(\theta) = \argmax  \mathbb{E}_{\tau \sim \pi_{\theta}(\tau)} [  R_0^T  ]
\end{equation} 

The action-value function describes what the expected return of the agent is in state $s$ and action $a$ under the policy $\pi$. The advantage of action value function is to make actions explicit, so we can select actions even in the model-free environment. After taking an action $a_t$ in state $s_t$ and thereafter following policy $\pi$, the action value function is formatted as:
\begin{align} \label{eq:q_value}
Q^{\pi}(s_t, a_t) = \mathbb{E}_{s_i\sim p_{\pi}, a_i\sim {\pi}} [R_{t}| s_t, a_t ] = \mathbb{E}_{s_i\sim p_{\pi}, a_i\sim {\pi}} [ \sum_{i=t}^T   \gamma^{(i-t)} r(  s_i, a_i   ) | s_t, a_t]
\end{align}

To get the optimal value function, we can use the maximum over actions, denoted as $Q^*(s_t, a_t) = \max_{\pi} Q^{\pi}(s_t, a_t)$, and the corresponding optimal policy $\pi$ can be easily derived
by $\pi^*(s) \in \argmax_{a_t} Q^*(s_t, a_t)$.

\subsection{Q-learning}
Suppose we use neural network parametrized by $\theta^{Q}$ to approximate Q-value in the continuous environment. To update Q-value function, we minimize the follow loss:
\begin{align}\label{eq:bellman}
L(\theta^{Q}) = \mathbb{E}_{s_i\sim p_{\pi}, a_i\sim {\pi}} [ ( Q(s_t, a_t ; \theta^{Q}) - y_t )^2   ]  
\end{align}
where $y_t = r(s_t, a_t) + \gamma \max_{a_{t+1}}  Q(s_{t+1}, a_{t+1} ; \theta^{Q}) $ is from Bellman equation, and its action $a_{t+1}$ is taken from frozen target networks (actor and critics) to stabilizing the learning.

In actor-critic methods, the policy $\pi: \mathcal{S} \mapsto \mathcal{A}$, known as the actor with parameters $\theta^{\pi}$, can be updated through the chain rule in the deterministic policy gradient algorithm \cite{Silver14}
\begin{align}\label{eq:policygrad}
\nabla J(  \theta^{\pi}   )  = \mathbb{E}_{s\sim p_{\pi}} [  \nabla_{a} Q(s, a; \theta^{Q} )|_{a=\pi(s; \theta^{\pi}   ) }   \nabla_{ \theta^{\pi} } ( \pi(s;  \theta^{\pi}   )  )   ]
\end{align}
where $Q(s, a)$ is the expected return while taking action $a$ in state $s$, and following $\pi$ after. One issue has been attracted great attention is overestimation bias, which may exacerbate the situation into a more significant bias over the following updates if left unchecked. Moreover, an inaccurate value estimate may lead to poor policy updates. 

Double Q-learning use two independent critics $q_1(s_t, a_t)$ and $q_2(s_t, a_t)$, where policy selection uses a different critic network than value estimation 
\begin{align}
q_1(s_t, a_t) = r(s_t, a_t) + \gamma q_2(s_{t+1},   \argmax_{a_{t+1}}  q_1(s_{t+1}, a_{t+1} ; \theta^{q_1})  ; \theta^{q_2} ) \nonumber \\
q_2(s_t, a_t) = r(s_t, a_t) + \gamma q_1(s_{t+1},   \argmax_{a_{t+1}}  q_2(s_{t+1}, a_{t+1} ; \theta^{q_2})  ; \theta^{q_1} ) \nonumber
\end{align}
Then the same square loss in Eq. \ref{eq:bellman} can be used to learn model parameters.

\section{Decorrelated Double Q-learning}
In this section, we introduce Decorrelated Double Q-learning (D2Q) for continuous action control. Similar to Double Q-learning, we use two value functions to approximate $Q(s_t, a_t) $. Our main contribution is to borrow the idea from control variates to decorrelate these two value functions, which can further reduce the overestimation risk. 
\subsection{Q-value function}
Suppose we have two approximators $q_1(s_t, a_t)$ and $q_2(s_t, a_t)$, D2Q uses the weighted difference of double q-functions to approximate the action-value function at $(s_t, a_t)$. Thus, we define Q-value as following:
\begin{align}\label{eq:dqlearning}
Q(s_t, a_t) = q_1(s_t, a_t) -  \beta \big( q_2(s_t, a_t) - E(q_2(s_t, a_t))  \big)
\end{align}
where $a_{t} = \pi (s_{t}; \theta^{\pi} )$, $q_2(s_t, a_t) - E(q_2(s_t, a_t))$ is to model the noise in state $s_t$ and action $a_t$, and $\beta \in(0, 1)$ is the correlation coefficient of $q_1(s_t, a_t)$ and $q_2(s_t, a_t)$. 
Thus, the weighted difference between $q_1(s_t, a_t)$ and $q_2(s_t, a_t)$ attempts to reduce the variance and remove the noise effects in Q-learning.  

To update $q_1$ and $q_2$, we minimize the following loss:
\begin{align}\label{eq:qdloss}
L(\theta^{Q}) &= \mathbb{E}_{s_i\sim p_{\pi}, a_i\sim {\pi}} [ ( q_1(s_t, a_t ; \theta^{q_1}) - y_t )^2   ]   + \mathbb{E}_{s_i\sim p_{\pi}, a_i\sim {\pi}} [ ( q_2(s_t, a_t ; \theta^{q_2}) - y_t )^2   ]   \nonumber \\
&+  \lambda  \mathbb{E}_{s_i\sim p_{\pi}, a_i\sim {\pi}} [ corr( q_1(s_t, a_t ; \theta^{q_1}),  q_2(s_t, a_t ; \theta^{q_2}) ) ] ^2
\end{align}
where $\theta^{Q} = \{ \theta^{q_1},  \theta^{q_2}   \}$, and $y_t $ is defined as 
\begin{align}\label{eq:tgtvalue}
y_t = r(s_t, a_t) +  \gamma  \textrm{min} (Q(s_{t+1}, a_{t+1}), q_2(s_{t+1}, a_{t+1}) )  
\end{align}
where $Q(s_{t+1}, a_{t+1})$ is the action-value function defined in Eq. \ref{eq:dqlearning} to decorrelate $q_1(s_{t+1}, a_{t+1})$ and $q_2(s_{t+1}, a_{t+1})$, which are both from the frozen target networks. To stabilize the target value, we take the minimum of $Q(s_{t+1}, a_{t+1})$ and $q_2(s_{t+1}, a_{t+1})$ in Eq. \ref{eq:tgtvalue} as TD3 \cite{FujimotoHM18}; and $corr( q_1(s_t, a_t ; \theta^{q_1}),  q_2(s_t, a_t ; \theta^{q_1}) ) $ measures similarity between these two value approximators. And the action $a_{t+1}$ is from policy $a_{t+1} = \pi (s_{t+1}; \theta^{\pi} ) $, which can take a similar policy gradient as in Eq. \ref{eq:policygrad}.  

Thus, the loss in Eq. \ref{eq:qdloss} tries to minimize the three terms below, as 
\begin{align}
& corr( q_1(s_t, a_t ; \theta^{q_1}),  q_2(s_t, a_t ; \theta^{q_2}) ) \to 0  \nonumber  \\ 
& q_1(s_t, a_t ; \theta^{q_1} )   \to y_t  \nonumber \\
 & q_2(s_t, a_t ; \theta^{q_2} ) \to y_t \nonumber
\end{align}

The purposes that we introduce control variate $q_2(s_t, a_t)$ are following:
(1) Since we use $q_2(s_t, a_t) - E(q_2(s_t, a_t)) $ to model noise, if there is no noise, such that $q_2(s_t, a_t) - E(q_2(s_t, a_t))= 0$, then we have $q_1(s_t, a_t) = Q^{\pi}(s_t, a_t)$, which is exactly the same as Double Q-learning. In experiment, we use the target value function $q_2(s_t, a_t)$ to approximate its expectation.
In fact, because of the noise in value estimate, we have $q_2(s_t, a_t) -E(q_2(s_t, a_t)) \neq 0$. The purpose we introduce $q_2(s_t, a_t)$ is to mitigate overestimate bias in Q-learning. 
(2) The control variate introduced by $q_2(s_t, a_t)$ will reduce the variance $Q(s_{t}, a_t )$ to stabilize the learning of value function.

There is existed theorem in \cite{Jaakkola94}, given the random process $\{\Delta_t \}$ taking value in $\mathbb{R}^n$ and defined as
\begin{align}
\Delta_{t+1} ( s_t,  a_t ) = (1- \alpha_t (s_t, a_t)  )  \Delta_{t} ( s_t,  a_t ) + \alpha_t(s_t, a_t) F_t(s_t, a_t)
\end{align}
converges to zero with probability 1 under the following assumptions:
\begin{enumerate}
 \item $0<\alpha_t <1, \sum_t \alpha_t(x) = \infty$ and  $\sum_t \alpha_t^2(x) <\infty $
\item $|| E[ F_t(x) | \mathcal{F}_t ] ||_{W} < \gamma || \Delta_t ||_{W}$ with $\gamma<1$
\item $var [  F_t(x) | \mathcal{F}_t   ]  \le C(1 + || \Delta_t ||^2_{W} )$ for $C>0$
 \end{enumerate}
 where $\mathcal{F}_t $ is a sequence of increasing $\sigma$-field such that $\alpha_t (s_t, a_t)$ and $\Delta_t$ are $\mathcal{F}_t $ measurable for $t=1,2,...$.
 
{\bf Convergence analysis}: we claim that our D2Q algorithm is convergence in the finite MDP settings. Based on the theorem above, we provide proof sketch below:
\begin{align}
\Delta_{t+1} ( s_t,  a_t ) &= (1- \alpha_t (s_t, a_t)  ) ( Q(s_t, a_t) - Q^*(s_t, a_t )  )   \nonumber \\
                                        &+ \alpha_t(s_t, a_t) \big(r_t + \gamma \min( Q(s_t, a_t ), q_2(s_t, a_t) ) - Q^*(s_t, a_t) \big) \nonumber \\
                                        & = (1- \alpha_t (s_t, a_t)  )  \Delta_{t} ( s_t,  a_t ) + \alpha_t(s_t, a_t) F_t(s_t, a_t)
\end{align}
where $F_t(s_t, a_t)$ is defined as:
\begin{align}
F_t(s_t, a_t) & = r_t + \gamma \min( Q(s_t, a_t ), q_2(s_t, a_t) ) - Q^*(s_t, a_t) \nonumber \\
                      & = r_t + \gamma \min( Q(s_t, a_t ), q_2(s_t, a_t) ) - Q^*(s_t, a_t) + \gamma Q(s_t, a_t) - \gamma Q(s_t, a_t) \nonumber \\
                      & = r_t + \gamma Q(s_t, a_t)- Q^*(s_t, a_t) + \gamma \min( Q(s_t, a_t ), q_2(s_t, a_t) ) - \gamma Q(s_t, a_t)
\end{align}
Then we need to prove $\min( Q(s_t, a_t ), q_2(s_t, a_t) ) -  Q(s_t, a_t)$ converges to 0 with probability 1.
\begin{align}
 &\min( Q(s_t, a_t ), q_2(s_t, a_t) ) -  Q(s_t, a_t) \nonumber \\
 =&  \min( Q(s_t, a_t ), q_2(s_t, a_t) ) - q_2(s_t, a_t )  +  q_2(s_t, a_t ) -  Q(s_t, a_t)  \nonumber \\
 =& \min( Q(s_t, a_t )-q_2(s_t, a_t  ), 0)  - ( Q(s_t, a_t )-q_2(s_t, a_t  )  ) \nonumber \\
 =& \min( q_1(s_t, a_t)-q_2(s_t, a_t  ) -\beta(q_2(s_t, a_t  ) - E(q_2(s_t, a_t  ) ) ), 0) \nonumber \\
  & +  q_1(s_t, a_t)-q_2(s_t, a_t  ) -\beta(q_2(s_t, a_t  ) - E(q_2(s_t, a_t  ) ) ) 
\end{align}

Suppose there exists very small $\delta_1$ and $\delta_2$, such that $| q_1(s_t, a_t) - q_2(s_t, a_t) | \le \delta_1 $ and $| q_2(s_t, a_t) - E( q_2(s_t, a_t) ) | \le \delta_2 $, then we have
\begin{align}
 &\min( Q(s_t, a_t ), q_2(s_t, a_t) ) -  Q(s_t, a_t) \nonumber \\
 \le & 2 ( | q_1(s_t, a_t)-q_2(s_t, a_t  )  | + \beta | q_2(s_t, a_t  ) - E(q_2(s_t, a_t  ) ) | ) \nonumber \\
 = & 2 ( \delta_1 + \beta \delta_2 ) < 4 \delta
\end{align}
where $\delta = \max (\delta_1,  \delta_2 )$. Note that $\exists \delta_1$, $| q_1(s_t, a_t) - q_2(s_t, a_t) | \le \delta_1 $ holds because $\Delta_{t}(q_1,q_2) = | q_1(s_t, a_t) - q_2(s_t, a_t) |$ converges to zero. According Eq. \ref{eq:qdloss}, both $q_1(s_t, a_t)$ and $q_2(s_t, a_t)$ are updated with following
\begin{align}
q_{t+1}(s_t, a_t) =  q_{t}(s_t, a_t) + \alpha_t(s_t, a_t )  (y_t-q_{t}(s_t, a_t) )
\end{align}
Then we have $\Delta_{t+1}(q_1,q_2) = \Delta_{t}(q_1,q_2) - \alpha_t(s_t, a_t ) \Delta_{t}(q_1,q_2) = (1 - \alpha_t(s_t, a_t ))\Delta_{t}(q_1,q_2)$ converges to 0 as the learning rate satisfies $0<\alpha_t(s_t, a_t )<1$.

\subsection{Correlation coefficient}
The purpose we introduce $corr( q_1(s_t, a_t ),  q_2(s_t, a_t ) )$ in Eq. \ref{eq:qdloss} is to reduce the correlation between two value approximators $q_1$ and $q_2$. In other words, we hope $q_1(s_t, a_t )$ and $q_2(s_t, a_t )$ to be as independent as possible. In this paper, we define $corr(q_1, q_2)$ as:
\begin{align}\label{eq:coveff}
corr(q_1(s_t, a_t ),  q_2(s_t, a_t ) )  = cosine( f_{q_1}(s_t, a_t),  f_{q_2}(s_t, a_t) ) \nonumber
\end{align}
where $cosine(a,b)$ is the cosine similarity between two vectors $a$ and $b$. $f_{q}(s_t, a_t)$ is the vector representation of the last hidden layer in the value approximator $q(s_t, a_t )$.

According to control variates, the optimal $\beta$ in Eq.\ref{eq:dqlearning} is:
\begin{align}
\beta = \frac{ cov( q_1(s_t, a_t), q_2(s_t, a_t ) )  }{ var(q_1(s_t, a_t) )  } \nonumber
\end{align}
where $cov$ is the symbol of covariance, and $var$ represents variance. Considering it is difficult to estimate $\beta$ in continuous action space, we take an approximation here. In addition, to reduce the number of hyper parameters, we set $\beta = corr(q_1(s_t, a_t ),  q_2(s_t, a_t ) )$ in Eq. \ref{eq:dqlearning} to approximate the correlation coefficient of $q_1(s_t, a_t )$ and $q_2(s_t, a_t )$ since it is hard to get covariance in the continuous action space.
%

\subsection{Algorithm}
We summarize our approach in Algorithm. \ref{alg:algd2q}. As Double Q-learning, we use the target networks with a slow updating rate to keep stability under temporal difference learning. Our contributions are two folder: (1) introduce the loss to minimize the correlation between two critics, which can make $q_1(s_t, a_t )$ and $q_2(s_t, a_t)$ as random as possible, and then effectively reduce the overestimation risk; (2) add control variates to reduce variance in the learning procedure. 
\begin{algorithm}[tb]
\caption{Decorrelated Double Q-learning}
\label{alg:algd2q}
\begin{algorithmic}
\STATE Initialize a pair of critic networks $q_1(s, a; \theta^{q_1})$, $q_2(s, a; \theta^{q_2})$ and actor $\pi(s ; \theta^{\pi})$ with weights $\theta^{Q} = \{ \theta^{q_1}, \theta^{q_2} \}$ and $\theta^{\pi}$
\STATE Initialize corresponding target networks for both critics and actor ${\theta^{Q}}^{\prime}  $ and ${\theta^{\pi}}^{\prime}$;
\STATE Initialize the total number of episodes $N$, batch size and the replay buffer $\textbf{\emph{R}}$
\STATE Initialize the coefficient $\lambda$ in Eq. \ref{eq:qdloss}
\STATE Initialize the updating rate $\tau$ for target networks
\FOR{episode = 1 {\bfseries to} $N$}
\STATE Receive initial observation state $s_0$ from the environment
\FOR{$t=0$ {\bfseries to} $T$}
\STATE Select action according to $a_t = \pi(s_t ; \theta^{\pi})$
\STATE Execute action $a_t$ and receive reward $r_t$, $done$, and further observe new state $s_{t+1}$
\STATE Push the tuple $(s_t, a_t, r_t, done, s_{t+1})$ into $\textbf{\emph{R}}$
\ENDFOR

\STATE Sample a batch of $D = {(s_t, a_t, r_t, done, s_{t+1})}$ from $\textbf{\emph{R}}$
\FOR{ $t$ =1 to length($D$) }
\STATE Compute $Q(s_t, a_t)$ with target critic networks according to Eq. \ref{eq:dqlearning}
\STATE Compute target value $y_t$ via Eq. \ref{eq:tgtvalue}
\STATE Update critics $q_1$ and $q_2$ by minimizing $ \mathcal{L}(\theta^Q)$ in Eq. \ref{eq:qdloss}
\STATE Update actor $a = \pi(s ; \theta^{\pi})$ by maximizing $Q(s_t, a_t)$ value in Eq. \ref{eq:dqlearning}
\ENDFOR

\STATE Update the target critics  ${\theta^{Q}}^{\prime}   = (1-\tau) {\theta^{Q}}^{\prime}     + \tau \theta^{Q} $
\STATE Update the target actor ${\theta^{\pi}}^{\prime}  = (1 - \tau) {\theta^{\pi}}^{\prime}  + \tau  \theta^{\pi} $ 
\ENDFOR
\STATE Return parameters $\theta = \{ \theta^{Q}, \theta^{\pi} \}$.
\end{algorithmic}
\end{algorithm}

\section{Experimental results}

In this section, we evaluate our method on the suite of MuJoCo continuous control tasks. We downloaded the OpenAI Gym environment, and used the v2 version of all tasks to test our method.

Without other specification, we use the same parameters below for all environments. The deep architecture for both actor and critic uses the same networks as TD3 \cite{FujimotoHM18}, with hidden layers [400, 300, 300]. Note that the actor adds the noise $\mathcal{N}(0, 0.1)$ to its action space to enhance exploration and the critic networks have two Q-functions $q_1(s, a)$ and $q_2(s, a)$. The mini-batch size is 100, and both network parameters are updated with Adam using the learning rate $10^{-3}$. In addition, we also use target networks (for both actor and critics) to improve the performance as in DDPG and TD3. The target policy is smoothed by adding Gaussian noise $\mathcal{N}(0, 0.2)$ as in TD3. We set the balance weight $\lambda= 2$. Both target networks are updated with $\tau = 0.005$. In addition, the off-policy algorithm uses the replay buffer $\textbf{\emph{R}}$ with size $10^6$ for all experiments.  

We run each task for 1 million time steps and evaluate it every 5000 time steps with no exploration noise. We repeat each task 5 times with random seeds and get its mean and standard deviation respectively. And we report our evaluation results by averaging the returns with window size 10.

\begin{figure*}[t!]
\begin{tabular}{ccc}
\includegraphics[trim=13mm 8.8mm 20mm 15mm, clip, width=4.3cm]{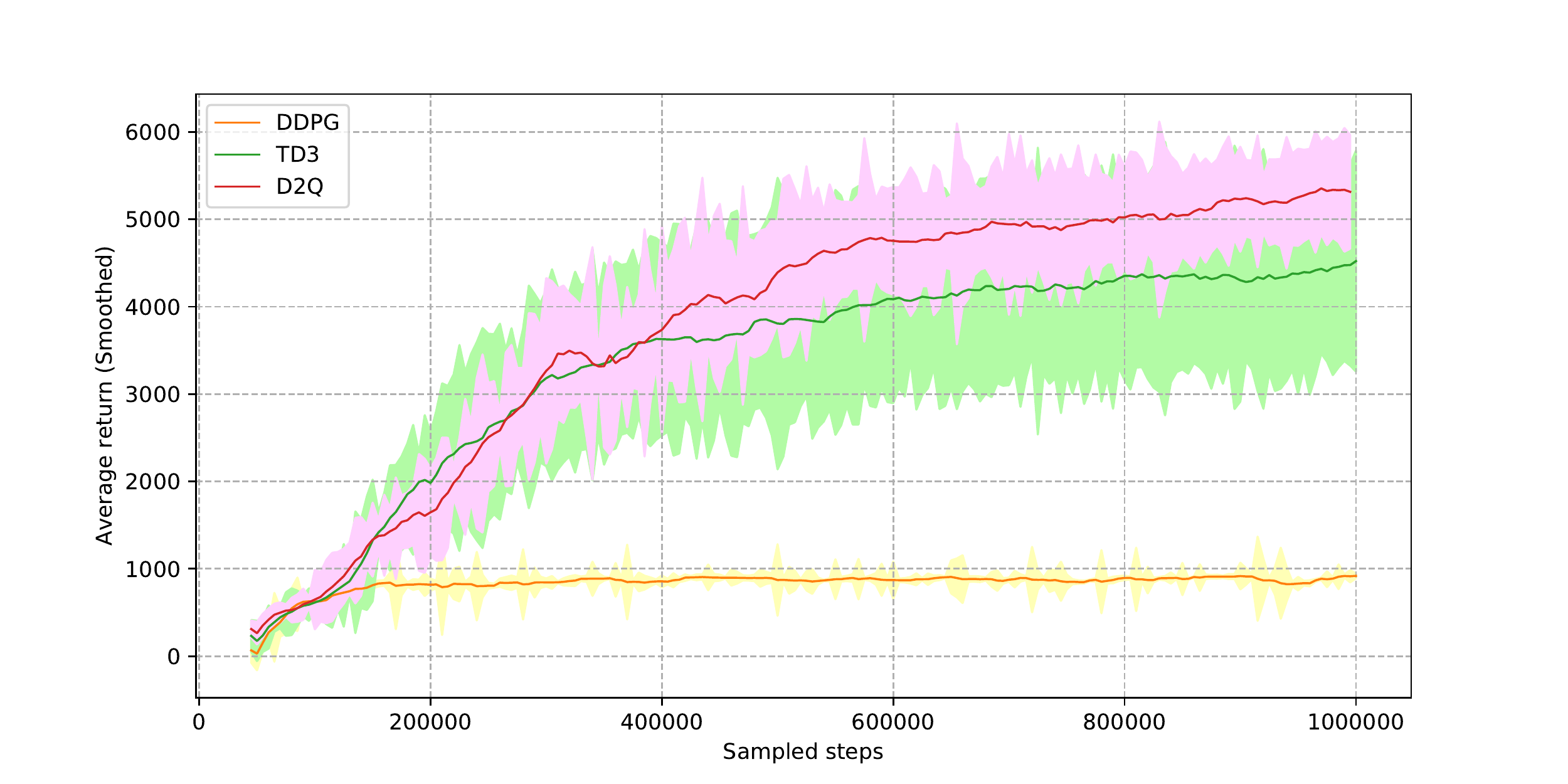} &
\includegraphics[trim=13mm 8.8mm 20mm 15mm, clip, width=4.3cm]{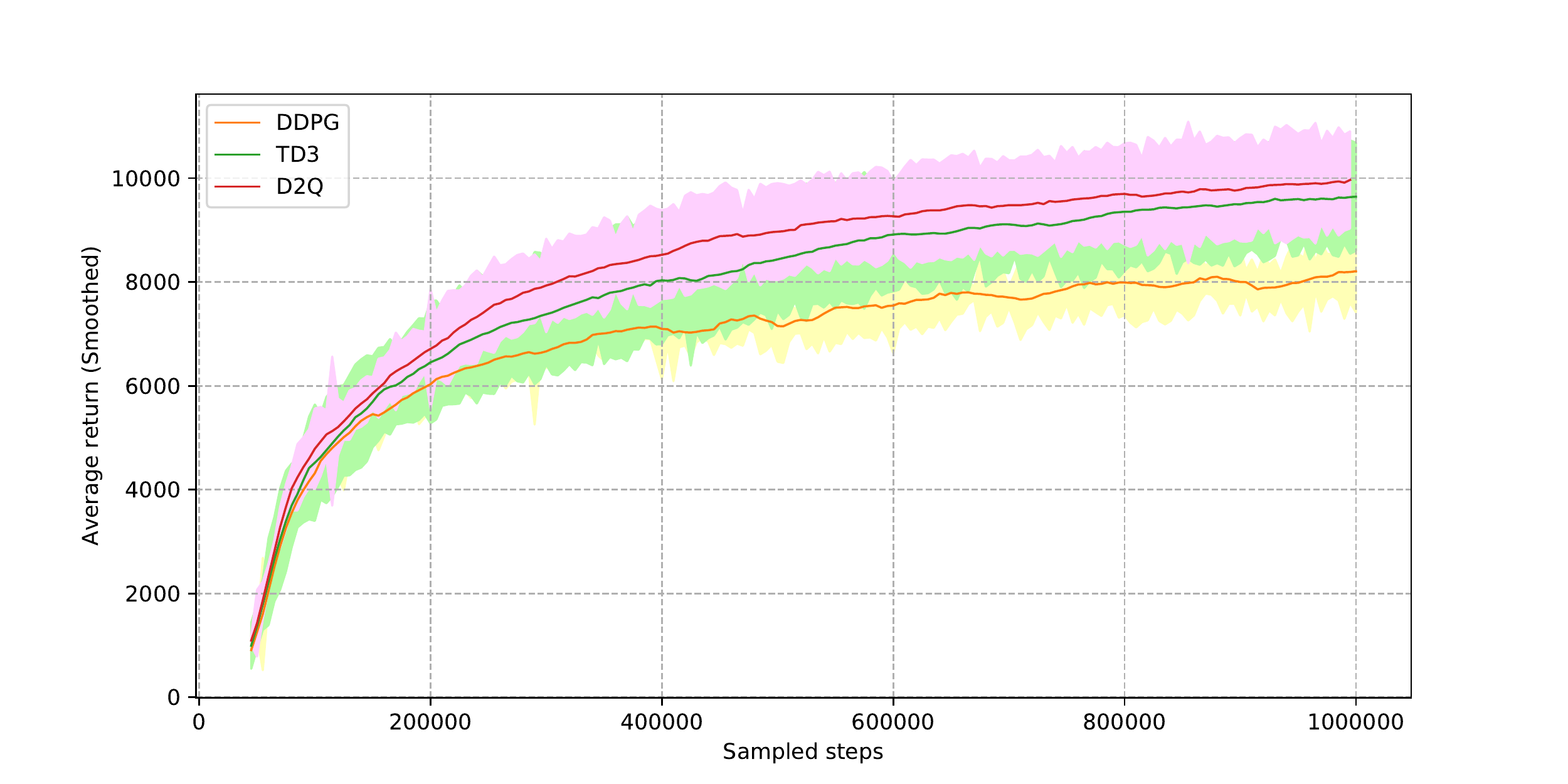} &
\includegraphics[trim=13mm 8.8mm 20mm 15mm, clip, width=4.3cm]{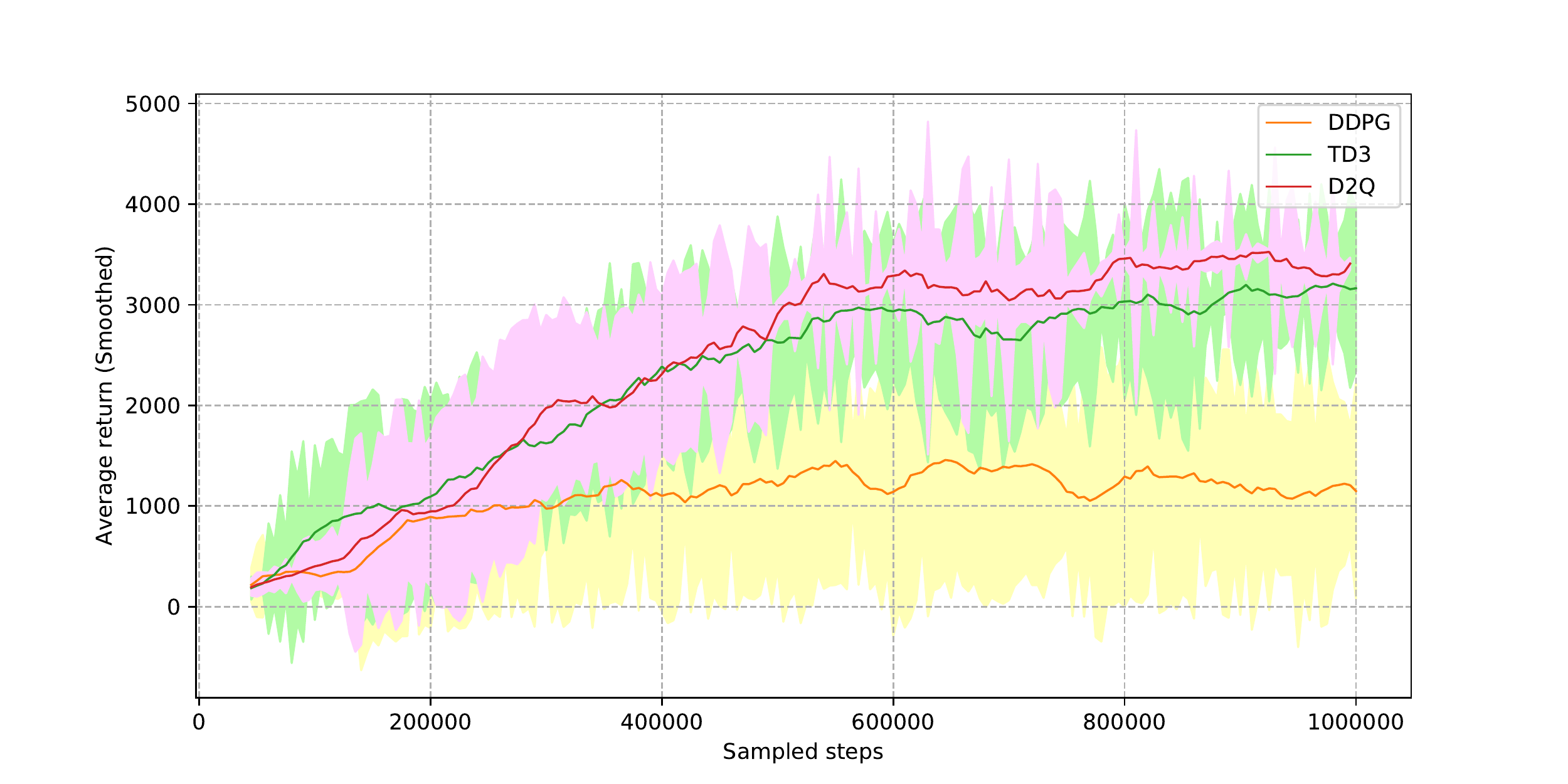} \\
(a) Ant & (b) Halfcheetah & (c) Hopper
\end{tabular}
\caption{The Learning curves with exploration noise on the Ant, Halfcheetah and Hopper environments. The shaded region represents the standard deviation of the average evaluation over nearby windows with size 10. Our D2Q algorithm yields significantly better results, compared to TD3 and DDPG.}
\label{Fig:MuJoCo1}
\end{figure*}

\begin{figure*}[t!]
\begin{tabular}{ccc}
\includegraphics[trim=13mm 8.8mm 20mm 15mm, clip, width=4.3cm]{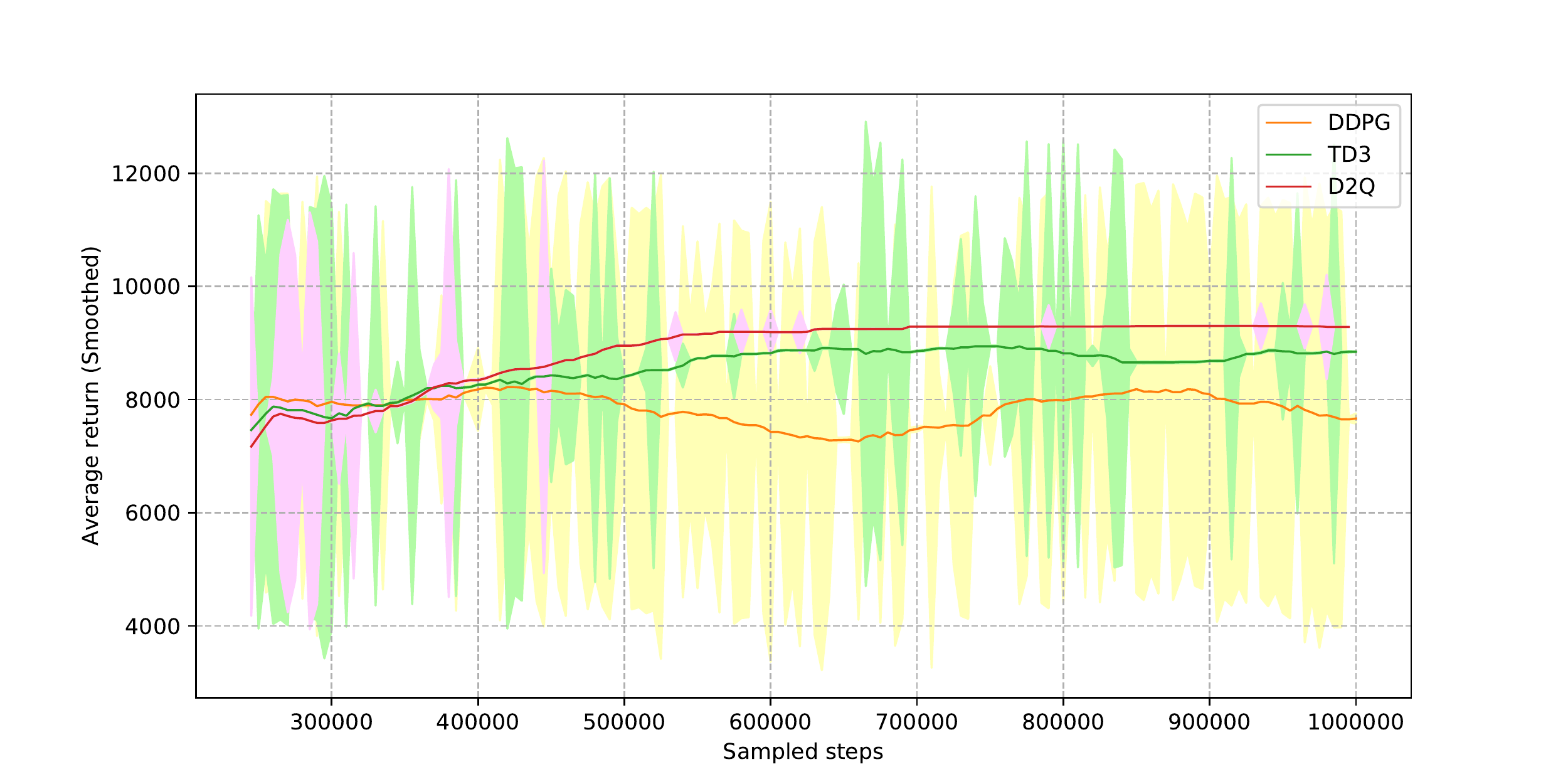} &
\includegraphics[trim=13mm 8.8mm 20mm 15mm, clip, width=4.3cm]{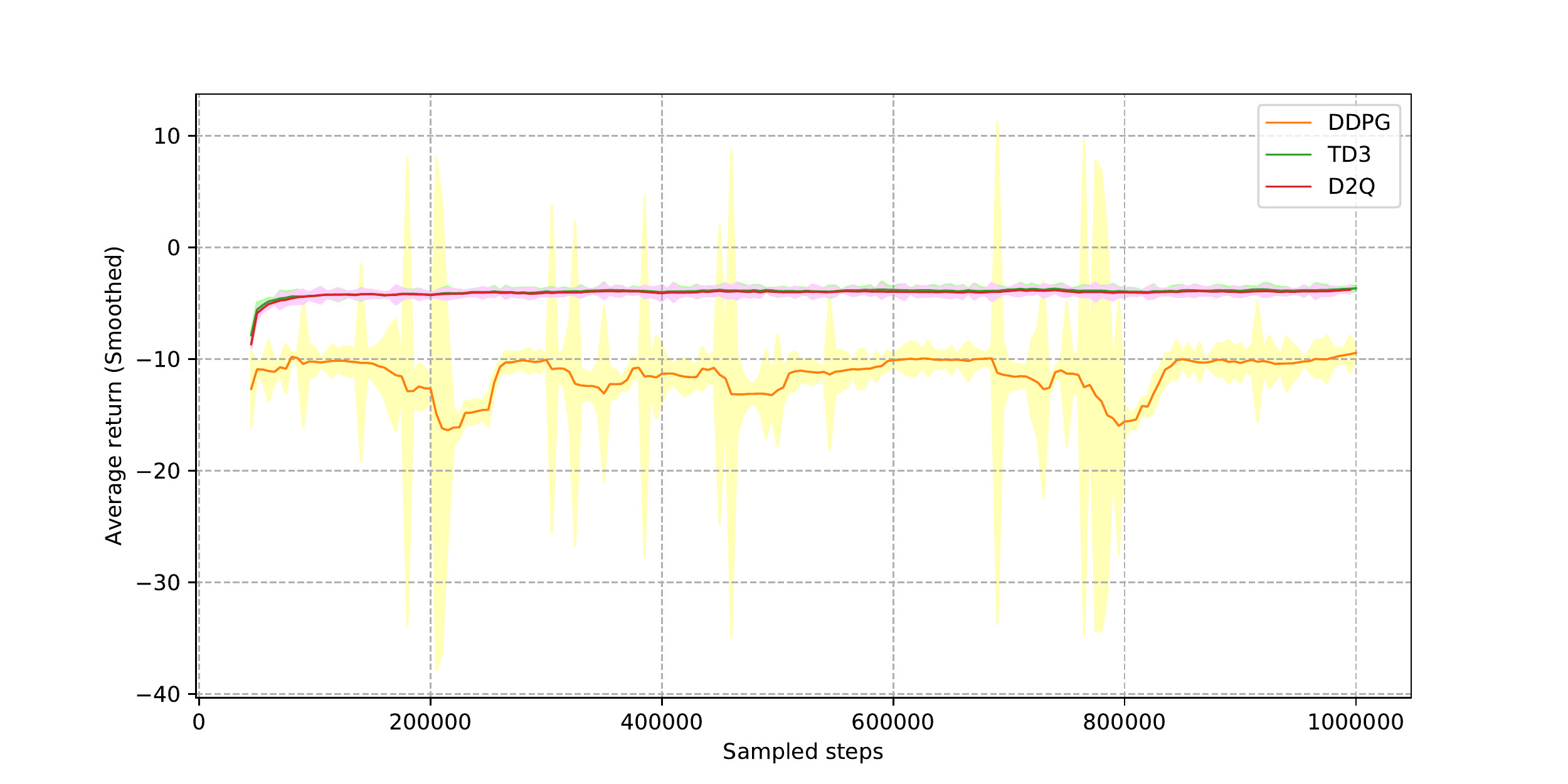} &
\includegraphics[trim=13mm 8.8mm 20mm 15mm, clip, width=4.3cm]{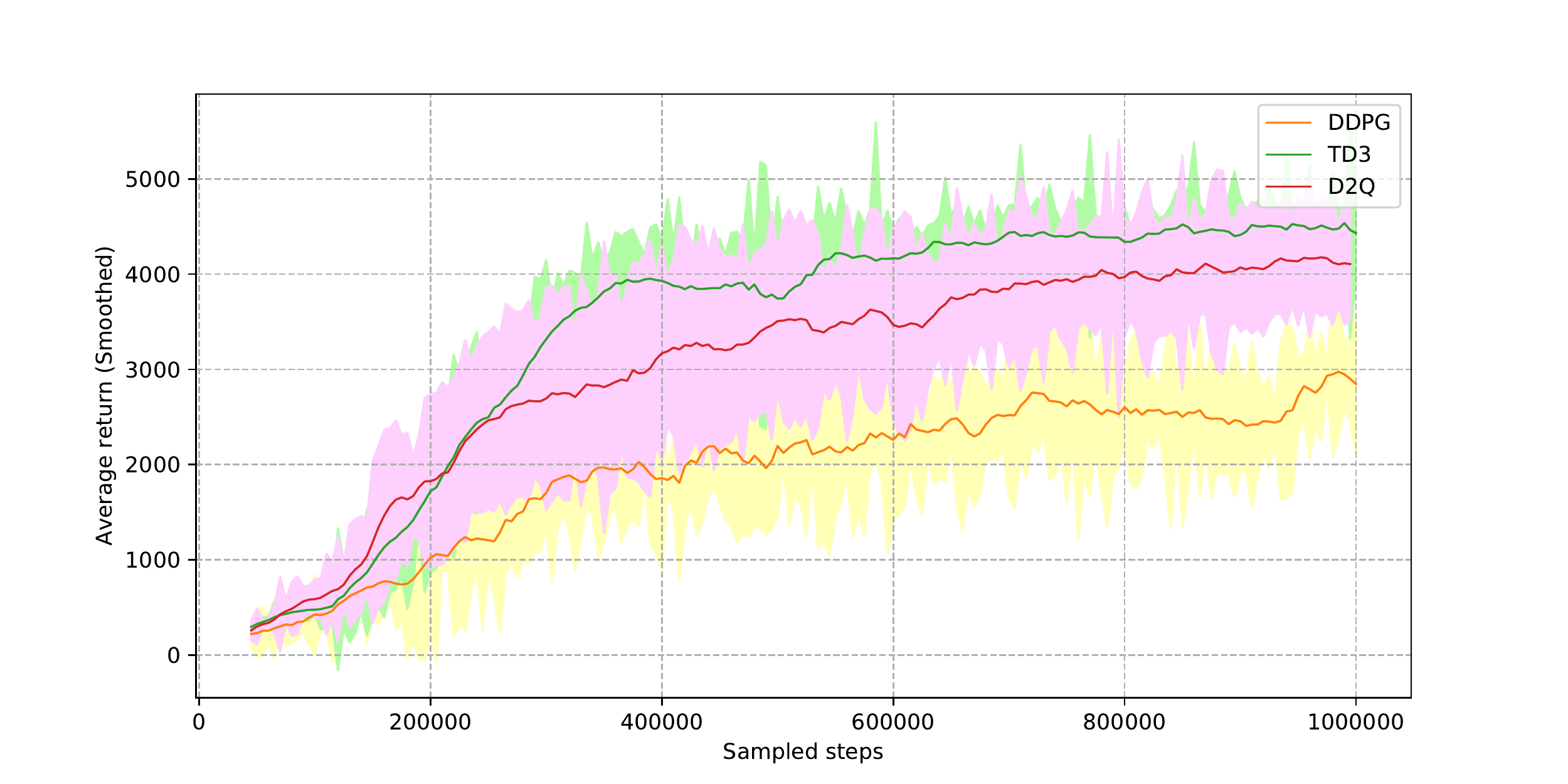} \\
(a) InvertedDoublePendulum & (b) Reacher & (c) Walker2d
\end{tabular}
\caption{The Learning curves with exploration noise on the InvertedDoublePendulum, Reacher and Walker2d environments. The shaded region represents the standard deviation of the average evaluation over nearby windows with size 10. Our D2Q algorithm yields competitive results, compared to TD3 and DDPG. Especially, our method gets better result on InvertedDoublePendulum task. }
\label{Fig:MuJoCo2}
\end{figure*}

The evaluation curves every 5000 time steps in the learning process are shown in Figures \ref{Fig:MuJoCo1} and \ref{Fig:MuJoCo2}. It demonstrates that our approach can yield competitive results, compared to TD3 and DDPG. Specifically, our D2Q method outperforms all other algorithms on Ant, HalfCheetah, Hopper and InvertedDoublePendulum. The quantitative results over 5 trials are presented in Table \ref{Tab:tab1}. Compared to SAC \cite{HaarnojaZAL18}, our approach shows better performance with lower variance given the same size of training samples. 

\begin{table*}[!h]
\caption{Comparison of Max Average Return over 5 trials of 1 million samples. The maximum value is marked bold for each task. $\pm$ corresponds to a single standard deviation over trials. }
\centering
\scalebox{0.62}{
\begin{tabular}{*8c}
\toprule
\multirow{2}{*}{Methods} &  \multicolumn{7}{c}{Environments} \\
{}   & HalfCheetah & Hopper  & Walker2d   & Ant &  Reacher & InvPendulum & InvDoublePendulum \\
\midrule
PPO   &  1795.43  & 2164.70   & 3317.69 & 1082.20 & -6.18 & 1000  & 8977.94\\
DDPG   &  8577.29  & 2020.46   & 1843.85 & 1005.30 & -6.51 & 1000  & $7741.28 \pm 2195.87$\\
TD3  &  $9636.95\pm859.06$ & $ 3223.75\pm514.2 $  & $\bf{4582.82\pm525.60}$  & $4373.44\pm1000.33$ & $ \bf{-3.6\pm0.56 }$ & $1000\pm0.0$  & $8911.04\pm750.58$\\
SAC\footnotemark[2] & $8895.96\pm 3316.5$ & $2100.67\pm1051.6$  & $3475.15\pm1508.71$  & $3250.49\pm 1157.94 $ & NA & NA  & NA\\
D2Q   &  $ \bf{9958.3\pm935.70} $ &  $\bf{3364.34 \pm 583.72}$    & $ 4129.20\pm270.71 $  & $\bf{5264.69\pm632.90}$ & $-3.78 \pm 0.32$ & $ \bf{1000 \pm 0.0} $ & $ \bf{9200.6\pm 186.22} $\\
\bottomrule
\end{tabular}}
\label{Tab:tab1}
\end{table*}


\section{Related work}
Considering the overestimation bias \cite{Thrun93} in Q-learning, many approaches have been proposed to avoid the maximization operator of a noisy value estimate. Delayed Q-learning \cite{Strehl06}  tries to find $\epsilon$-optimal policy, which determines how frequent to update state-action function. However, it can suffer from overestimation bias, although it guarantees to converge in polynomial time. Double Q-learning \cite{Hasselt10} introduces two independently trained critics to mitigate the overestimation effect. Averaged-DQN \cite{AnschelBS17} takes the average of previously learned Q-values estimates, which results in a more stable training procedure, as well as reduces approximation error variance in the target values. A clipped Double Q-learning called TD3 \cite{FujimotoHM18} extends the deterministic policy gradient \cite{Silver14,LillicrapHPHETS15} to address overestimation bias. In particular, TD3 uses the minimum of two independent critics to approximate the value function suffering from overestimation. Soft actor critic \cite{HaarnojaZAL18} takes a similar approach as TD3, but with better exploration with maximum entropy method. Maxmin Q-learning \cite{Lan20} extends Double Q-learning and TD3 to multiple critics to handle overestimation bias and variance. 

Another side effect of consistent overestimation \cite{Thrun93} in Q-learning is that the accumulated error of temporal difference \cite{SuttonB98} can cause high variance. To reduce the variance, there are two popular approaches: baseline and actor-critic methods \cite{Witten77,Konda00}. In policy gradient, we can minus baseline in Q-value function to reduce variance without bias. Further, the advantage actor-critic ($A^2C$) \cite{Mnih2016} introduces the average value to each state, and leverages the difference between value function and the average to update the policy parameters. Schulman et al proposed the generalized advantage value estimation \cite{Schulmanetal2016}, which considered the whole episode with an exponentially-weighted estimator of the advantage function that is analogous to $TD(\lambda)$  to substantially reduce the variance of policy gradient estimates at the cost of some bias.

From another point of view, baseline and actor-critic methods can be categories into control variate methods \cite{Greensmith01}. Greensmith et al. analyze the two additive control variate methods theoretically including baseline and actor-critic method to reduce the variance of performance
gradient estimates in reinforcement learning problems. 
Interpolated policy gradient (IPG) \cite{Gu17} based on control variate methods merges on- and off-policy updates to reduce variance for deep reinforcement learning. 
Motivated by the Stein's identity, Liu et al. introduce more flexible and general action-dependent baseline functions \cite{LiuFMZ0018} by extending the previous control variate methods used in REINFORCE and advantage actor-critic.

\section{Conclusion}
In this paper, we propose a Decorrelated Double Q-learning approach for off-policy reinforcement learning. We use a pair of critics for value estimate, but we introduce an objective function to decorrelated these two approixmators. In addition, we consider to leverage control variates to reduce variance and stabilize the learning procedure. The experimental results on a suite of Gym environment demonstrate our approach yields good performance than competitive baselines. 
%
%

\bibliographystyle{unsrt}
\bibliography{acpaper2019}

\begin{thebibliography}{10}

\bibitem{Watkins92}
Christopher J. C.~H. Watkins and Peter Dayan.
\newblock Q-learning.
\newblock In {\em Machine Learning}, pages 279--292, 1992.

\bibitem{Mnih13}
Volodymyr Mnih, Koray Kavukcuoglu, David Silver, Alex Graves, Ioannis
  Antonoglou, Daan Wierstra, and Martin Riedmiller.
\newblock Playing atari with deep reinforcement learning.
\newblock In {\em NIPS Deep Learning Workshop}. 2013.

\bibitem{Witten77}
Ian~H. Witten.
\newblock An adaptive optimal controller for discrete-time markov environments.
\newblock {\em Information and Control}, pages 286--295, 1977.

\bibitem{Konda00}
Vijay~R. Konda and John~N. Tsitsiklis.
\newblock Actor-critic algorithms.
\newblock In {\em Advances in Neural Information Processing Systems}, pages
  1008--1014. MIT Press, 1999.

\bibitem{Silver14}
David Silver and Guy Lever.
\newblock Deterministic policy gradient algorithms.
\newblock In {\em ICML}, 2014.

\bibitem{LillicrapHPHETS15}
Timothy~P. Lillicrap, Jonathan~J. Hunt, Alexander Pritzel, Nicolas Heess, Tom
  Erez, Yuval Tassa, David Silver, and Daan Wierstra.
\newblock Continuous control with deep reinforcement learning.
\newblock {\em CoRR}, abs/1509.02971, 2015.

\bibitem{Thrun93}
Sebastian Thrun and Anton Schwartz.
\newblock Issues in using function approximation for reinforcement learning.
\newblock In Michael Mozer, Paul Smolensky, David Touretzky, Jeffrey Elman, and
  Andreas Weigend, editors, {\em Proceedings of the 1993 Connectionist Models
  Summer School}, pages 255--263. Lawrence Erlbaum, 1993.

\bibitem{SuttonB98}
Richard~S. Sutton and Andrew~G. Barto.
\newblock {\em Reinforcement learning - an introduction}.
\newblock Adaptive computation and machine learning. {MIT} Press, 1998.

\bibitem{Hasselt10}
Hado van Hasselt.
\newblock Double q-learning.
\newblock In {\em Advances in Neural Information Processing Systems 23: 24th
  Annual Conference on Neural Information Processing Systems 2010. Proceedings
  of a meeting held 6-9 December 2010, Vancouver, British Columbia, Canada},
  pages 2613--2621. Curran Associates, Inc., 2010.

\bibitem{AnschelBS17}
Oron Anschel, Nir Baram, and Nahum Shimkin.
\newblock Averaged-dqn: Variance reduction and stabilization for deep
  reinforcement learning.
\newblock In {\em Proceedings of the 34th International Conference on Machine
  Learning, {ICML} 2017, Sydney, NSW, Australia, 6-11 August 2017}, volume~70
  of {\em Proceedings of Machine Learning Research}, pages 176--185. {PMLR},
  2017.

\bibitem{FujimotoHM18}
Scott Fujimoto, Herke van Hoof, and David Meger.
\newblock Addressing function approximation error in actor-critic methods.
\newblock In {\em ICML}, volume~80 of {\em {JMLR} Workshop and Conference
  Proceedings}, pages 1582--1591. JMLR.org, 2018.

\bibitem{HaarnojaZAL18}
Tuomas Haarnoja, Aurick Zhou, Pieter Abbeel, and Sergey Levine.
\newblock Soft actor-critic: Off-policy maximum entropy deep reinforcement
  learning with a stochastic actor.
\newblock In {\em {ICML}}, volume~80 of {\em {JMLR} Workshop and Conference
  Proceedings}, pages 1856--1865. JMLR.org, 2018.

\bibitem{Lan20}
Qingfeng Lan, Yangchen Pan, Alona Fyshe, and Martha White.
\newblock Maxmin q-learning: Controlling the estimation bias of q-learning.
\newblock In {\em ICLR}, 2020.

\bibitem{Jaakkola94}
T.~{Jaakkola}, M.~I. {Jordan}, and S.~P. {Singh}.
\newblock On the convergence of stochastic iterative dynamic programming
  algorithms.
\newblock {\em Neural Computation}, 6(6):1185--1201, 1994.

\bibitem{Strehl06}
Alexander~L. Strehl, Lihong Li, Eric Wiewiora, John Langford, and Michael~L.
  Littman.
\newblock Pac model-free reinforcement learning.
\newblock In {\em In: ICML-06: Proceedings of the 23rd international conference
  on Machine learning}, pages 881--888, 2006.

\bibitem{Mnih2016}
Volodymyr Mnih, Adri\`{a}~Puigdom\`{e}nech Badia, Mehdi Mirza, Alex Graves, Tim
  Harley, Timothy~P. Lillicrap, David Silver, and Koray Kavukcuoglu.
\newblock Asynchronous methods for deep reinforcement learning.
\newblock In {\em Proceedings of the 33rd International Conference on
  International Conference on Machine Learning - Volume 48}, ICML'16, pages
  1928--1937. JMLR.org, 2016.

\bibitem{Schulmanetal2016}
John Schulman, Philipp Moritz, Sergey Levine, Michael Jordan, and Pieter
  Abbeel.
\newblock High-dimensional continuous control using generalized advantage
  estimation.
\newblock In {\em Proceedings of the International Conference on Learning
  Representations (ICLR)}, 2016.

\bibitem{Greensmith01}
Evan Greensmith, Peter~L. Bartlett, and Jonathan Baxter.
\newblock Variance reduction techniques for gradient estimates in reinforcement
  learning.
\newblock In {\em Journal of Machine Learning Research}, pages 1471--1530. MIT
  Press. In press, 2001.

\bibitem{Gu17}
S.~Gu, T.~Lillicrap, R.~E. Turner, Z.~Ghahramani, B.~Sch{\"o}lkopf, and
  S.~Levine.
\newblock Interpolated policy gradient: Merging on-policy and off-policy
  gradient estimation for deep reinforcement learning.
\newblock In {\em Advances in Neural Information Processing Systems 30}, pages
  3849--3858. Curran Associates, Inc., 2017.

\bibitem{LiuFMZ0018}
Hao Liu, Yihao Feng, Yi~Mao, Dengyong Zhou, Jian Peng, and Qiang Liu.
\newblock Action-dependent control variates for policy optimization via stein
  identity.
\newblock In {\em 6th International Conference on Learning Representations,
  {ICLR} 2018, Vancouver, BC, Canada, April 30 - May 3, 2018, Conference Track
  Proceedings}. OpenReview.net, 2018.

\end{thebibliography}

\end{document}